\documentclass[a4paper, 10pt, conference]{ieeeconf}      

\usepackage{graphicx}

\usepackage{epsfig}
\usepackage{amsmath, amssymb}
\usepackage{algorithmic}
\usepackage{algorithm}
\usepackage{url}

\IEEEoverridecommandlockouts                              
\title{\LARGE \bf
Particle Filter Based Monocular Human Tracking with a 3D Cardbox Model and a Novel Deterministic Resampling Strategy}

\author{\parbox{5 in}{\centering Ziyuan Liu $^\dag$,
 Dongheui Lee $^\dag$ and Wolfgang Sepp $^\ddag$\\
              $^\dag$ Institute of Automatic Control Engineering, Technische Universit\"at M\"unchen, D-80290, M\"unchen, Germany\\
              $^\ddag$ Institute of Robotics and Mechatronics, German Aerospace Center (DLR), D-82234, Wessling, Germany.\\
               \small{\texttt{\{ziyuan.liu, dhlee\}@tum.de, wolfgang.sepp@dlr.de}}
               }
}

\begin{document}

\maketitle
\thispagestyle{empty}
\pagestyle{empty}

\begin{abstract}
The challenge of markerless human motion tracking is the high dimensionality of the search space. Thus, efficient exploration in the search space is of great significance. In this paper, a motion capturing algorithm is proposed for upper body motion tracking. The proposed system tracks human motion based on monocular silhouette-matching, and it is built on the top of a hierarchical particle filter, within which a novel deterministic resampling strategy (DRS) is applied. The proposed system is evaluated quantitatively with the ground truth data measured by an inertial sensor system. In addition, we compare the DRS with the stratified resampling strategy (SRS). It is shown in experiments that DRS outperforms SRS with the same amount of particles. Moreover, a new 3D articulated human upper body model with the name 3D cardbox model is created and is proven to work successfully for motion tracking. Experiments show that the proposed system can robustly track upper body motion without self-occlusion. Motions towards the camera can also be well tracked.
\end{abstract}

\section{INTRODUCTION}
Human motion capture is the process in which the configuration of body parts is estimated over time from sensor input. Marker-based human motion capture systems use markers attached to the human body to capture human motion. Despite the good accuracy of marker-based motion capture systems, they are obtrusive and expensive. Many applications, such as surveillance and human-computer interaction applications, however, require the capture system to be markerless, and to capture human motion just by analyzing image sequences. In recent years, these systems have attracted tremendous attention, and they are considered to be an active research area in the future as well.\\
\indent In this paper, a motion capture system is proposed for upper body motion tracking. The proposed system tracks human motion based on monocular silhouette-matching, and it is built on the basis of a hierarchical particle filter. In order to tackle the high dimensionality of the search space, an efficient \emph{deterministic resampling strategy} (DRS) is applied within the particle filter framework. Using the ground truth data which is obtained by the Moven inertial motion capture suit \cite{04}, we evaluate the proposed system quantitatively. In experiments, we compare our DRS with the stratified resampling strategy (SRS) \cite{16}. Experiments show that the proposed system achieves stable real-time tracking of human motion without self-occlusion in 15 fps (frames per second) using 600 particles. Motions towards the camera can also be well tracked. Additionally, a new 3D articulated human upper body model with the name \textit{3D cardbox model} is created and is proven to work successfully for motion tracking. \\
\indent The reminder of this paper is structured as follows: in Section \ref{0}, we review the related work on markerless human motion tracking and human modeling. Subsequently, we describe our system with respect to its two parts: modeling and estimation. In Section \ref{2}, we introduce the modeling part, where our new human model and other basic components of the modeling part are explained. In Section \ref{3}, the estimation part of the proposed system is introduced, where we discuss model initialization and online motion tracking respectively. In Section \ref{4}, the proposed system is evaluated in experiments, where we compare our DRS with SRS.
\section{RELATED WORK}\label{0}
\indent The main challenge of markerless human motion capturing lies at the high dimensionality of the search space. So far, many approaches have been proposed. Particle filtering \cite{1-1} samples the configuration space according to stochastic models of the observation. However, the bottleneck of applying particle filtering in markerless motion tracking is that the number of particles needed for a successful tracking increases exponentially with the number of degrees of freedom (DOF) of the underlying human model. According to \cite{1}, a realistic human model contains 14 DOF for upper body or 25 DOF for full body, and this results in an extremely high-dimensional search space. In order to address this problem, the annealed particle filter \cite{4} and the partitioned particle filter \cite {4-1} have been proposed, in which the number of particles needed is effectively reduced, however, the processing speed is slow. Another way to solve the high-dimensionality is to add constraints to human movements, as done in \cite {4-2}. Nevertheless, the power of such systems is limited due to the incorporated constraints. L. Sigal et. al. \cite{5} apply optimized algorithms to search for a global or local estimation of the pose configuration, but the applicability of the system is limited to situations, in which multiple cameras are used. Gall et. al. \cite{7-1} combine the stochastic sampling and the optimization to search for a global estimate of the human posture. Despite the good tracking accuracy, their system needs no less than 76 seconds to process one frame. Other systems are based on additional 3D sensor, such as the Swiss Ranger, and employ Iterative Closest Point (ICP) algorithm \cite{14}. Recently, systems using the Kinect sensor have also been proposed \cite{15}. As pointed out by \cite{15-1}, one limitation for most of the existing systems is the lack of quantitative error analysis, which is important for system comparison. \\
\indent In addition to different tracking methods, there are also several ways to model human body. According to \cite{00}, there exist nowadays three main types of human models: 2D shape model \cite{24}, 3D volumetric model \cite{33} and 3D surface-based shape model \cite{34}. In 2D models, body parts are modeled as 2D patches. Such models usually do not represent the geometry and kinematics of the human body explicitly. Normally, scaling factors are calculated for 2D models, so that they can better match the input images. In contrast to 2D models, 3D models incorporate physical rotations and translations of the body parts to produce the final body postures. In 3D volumetric models, body parts are modeled as separate rigid objects that are articulated through joints. In 3D surface-based models, the human body is modeled as a single surface which is built by a mesh of deformable polygons.




\section{MODELING}\label{2}
\indent The modeling part of the proposed system contains two components: physical model and observation model. They are explained in the following.
\subsection{Physical Model}
\begin{figure}[tp]
   \centering
      \includegraphics[height=3.3cm]{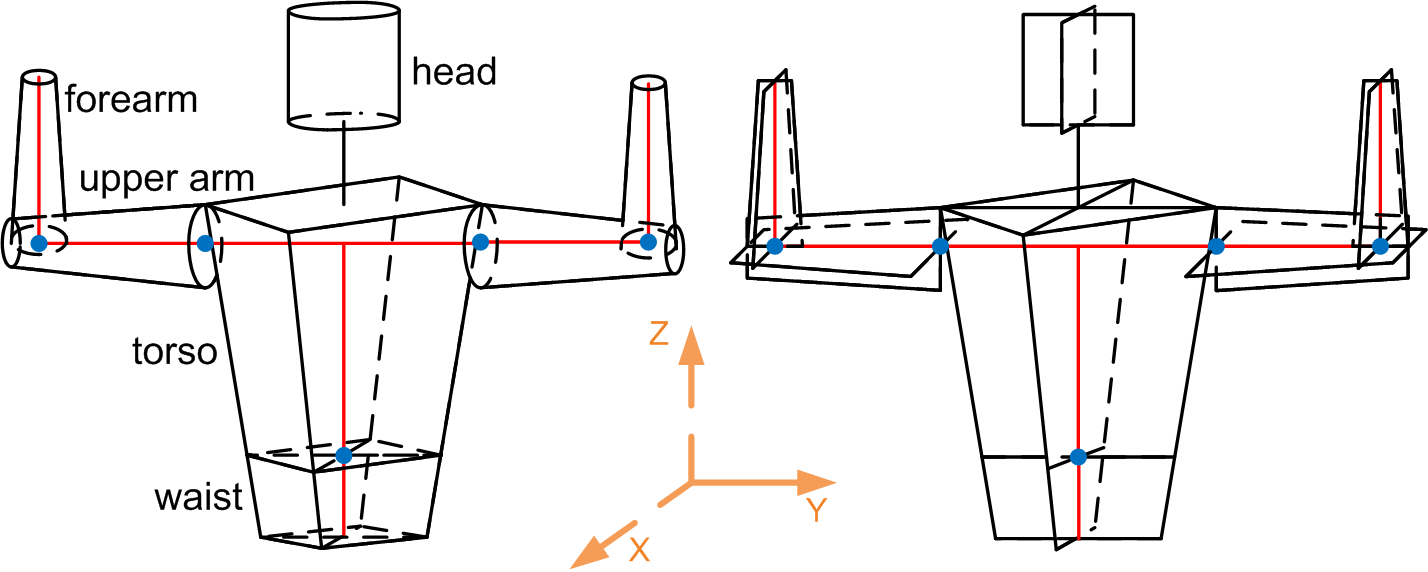}
    \caption{The 3D volumetric model (left) and the 3D cardbox model (right). Red lines denote the skeleton and blue circles denote joints. Orange arrows (X, Y and Z) show the 3 DOF for determining the torso position, two out of which are assumed constant (orange dashed arrows: X and Z). Axis X is the opposite direction of the optical axis of the camera. Axis Y is horizontally parallel to the camera plane, and Axis Z is determined by the right-hand rule.}
   \label{figure:comparison}
   \end{figure}

\begin{figure}[tp]
   \centering
      \includegraphics[height=5cm]{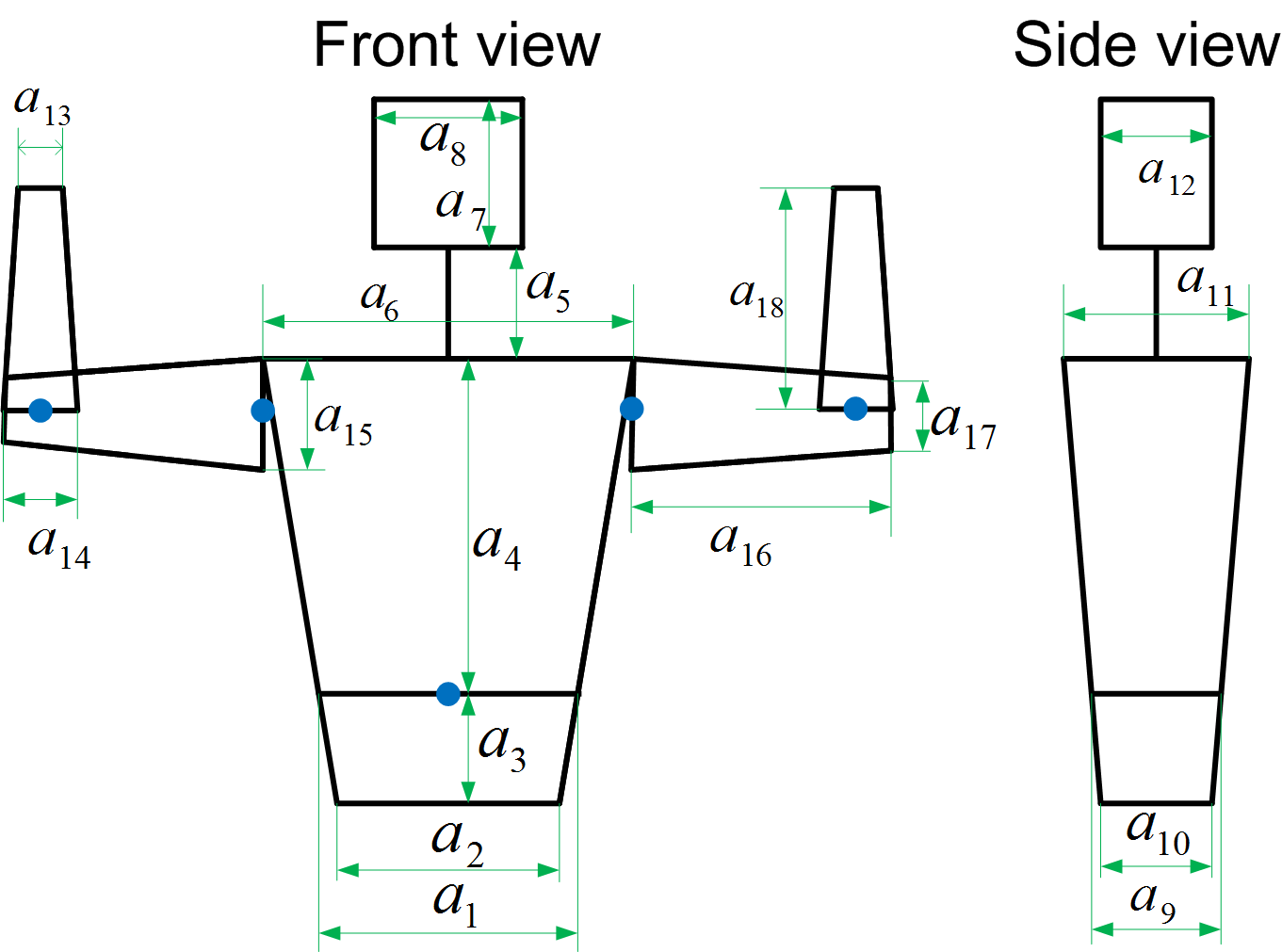}
    \caption{Size parameters of the 3D cardbox model.}
   \label{figure:size}
   \end{figure}

\indent In this paper, a new 3D articulated human model with the name ``3D cardbox model'' is proposed. Our 3D cardbox model represents each body part with two or three 2D patches that are perpendicular to each other. A comparison of the 3D volumetric model and our 3D cardbox model is illustrated in Fig. \ref{figure:comparison}. In the 3D cardbox model, the head, upper arm, forearm and waist are represented by two central patches, whereas the torso is modeled by three patches. Compared with the 3D volumetric model, our model has less planes to project while rendering the human model image.\\
\indent Our human model contains two types of parameters: \textit{posture parameters} and \textit{size parameters}. The posture parameters refer to the kinematic model of our 3D cardbox model, which contains 12 DOF in all, these are: two DOF for rotations of each elbow, two DOF for rotations of each shoulder, three DOF for the orientation of the torso and one DOF for the position of the torso. Intuitively, three DOF are needed for determining torso position, however, at the current stage, two of them are used as constant after an initialization step. These two DOF are along the directions marked by dashed orange arrows in Fig. \ref{figure:comparison}. Thus, we estimate 12 DOF during online motion tracking. Given all the DOF information, the 3D model is mapped to the image plane of a camera observing the scene. Size parameters of our human model reflect the physical dimension of the observed person. These parameters are initialized through offline model initialization. Fig. \ref{figure:size} illustrates all size parameters of our human model.

\subsection{Observation Model}
\begin{figure}[tp]
   \centering
      \includegraphics[height=3cm]{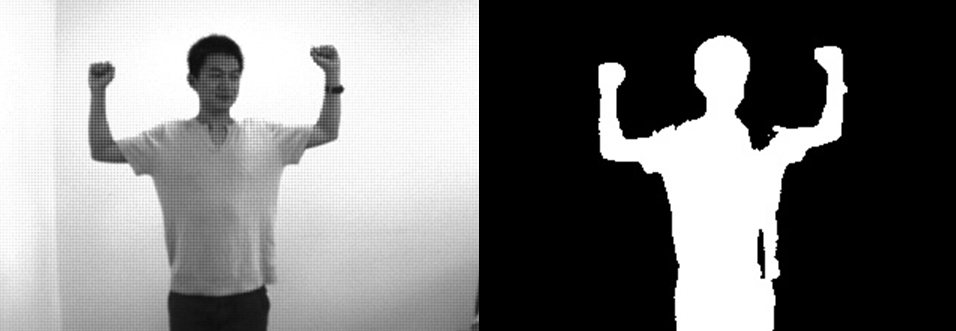}
    \caption{Foreground segmentation: observation image (left) and extracted silhouette image (right).}
   \label{figure:background}
   \end{figure}

\begin{figure}[bp]
   \centering
      \includegraphics[height=1.9cm]{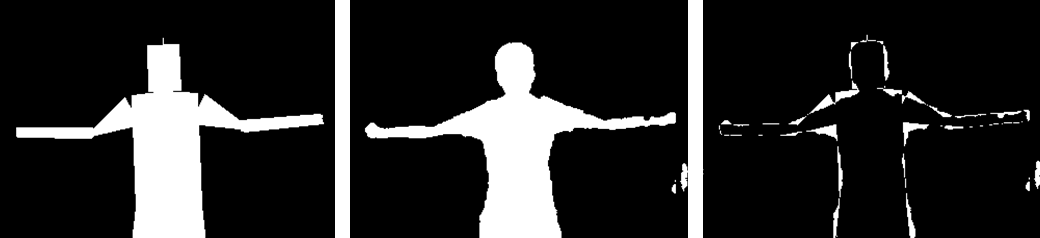}
    \caption{Silhouette matching: projected human model $h(x,y,\bold{d})$ (left), extracted silhouette $f(x,y,I_t)$ (middle) and distinct pixels (cost) $W(\bold{d},I_t)$ shown in white (right).}
   \label{figure:silhouette-matching}
   \end{figure}

\indent In our approach, we observe human motion by the silhouette of the person. Silhouette images are obtained via foreground-background segmentation. Here, the background and the camera position are both assumed static. The brightness statistics of the background is obtained by capturing first a set of images of the background and then using the averaging background method \cite{03}. In this way, the mean brightness $I_\textrm{B}(x,y)$ and standard deviation $\sigma_\textrm{B}(x,y)$ of the background is learned. Having built the background model, pixels of subsequent images $I_t(x,y)$, whose brightness is different from that of the learned background, are identified as foreground. The input image is thresholded into a binary silhouette image:
$$f(x,y,I_t) = $$
\begin{equation}\label{equ0001}
\left\{\begin{array}{cl}
 0, &\left\vert I_\textrm{B}(x,y)-I_t(x,y)\right\vert <2\sigma_\textrm{B}(x,y)\\
1, &\text{otherwise}
\end{array}
\right.
\end{equation}
where $0$ indicates the background and $1$ the foreground. Fig. \ref{figure:background} shows one example of foreground segmentation.\\
\indent A certain human posture $\bold{d}$ is assigned a cost $W(\bold{d},I_t)$ by matching the perspectively projected 3D model $h(x,y,\bold{d})$ with the silhouette image $f(x,y,I_t)$. The cost denotes the amount of distinct pixels between $h(x,y,\bold{d})$ and $f(x,y,I_t)$ and is given as:
\begin{eqnarray}\label{equ100}
W(\bold{d},I_t)=\sum_{x,y}h(x,y,\bold{d})\oplus f(x,y,I_t),\nonumber \\
\text{with}~h(x,y,\bold{d})=
\left\{\begin{array}{cl}
 1, &(x,y)\in D\\
0, &\text{otherwise}
\end{array}
\right.
\end{eqnarray}
where $D$ denotes the set of image coordinates that are located within the projected human model. The symbol $\oplus$ denotes the logic XOR function. The function $h(x,y,\bold{d})$ indicates the intensity value of pixel $(x,y)$ of the projected human model that is generated by the set of posture parameters $\bold{d}$:
\begin{equation}
\bold{d}=[d_1,d_2,\cdots,d_{12}]^T,
\end{equation}
where each element $d$ in $\bold{d}$ refers to one DOF. Fig. \ref{figure:silhouette-matching} shows one example of silhouette matching, where the distinct pixels (right) between the projected human model (left) and the extracted silhouette (middle) are shown in white. In Section \ref{3}, we explain how to estimate posture parameters and size parameters by minimizing the cost function in (\ref{equ100}).
\section{ESTIMATION}\label{3}
\indent The estimation part of the proposed system contains two phases, i.e. offline model initialization and online motion tracking as described in the following.
\subsection{Offline Model Initialization}
\indent Before the human model is used for online motion tracking, it must be initialized, which means the body size parameters of the human model are tuned for the observed person. Fig. \ref{figure:size} illustrates the body size parameters that need to be initialized. These parameters are iteratively estimated by local optimization around an initial guess using multiple hypotheses. A hypothesis of size parameters $\bold{A}$ is generated by adding Gaussian noise $\mathcal{N}(0,\sigma_{\alpha}^2E)$ to initial guess $\bold{A}_{ini}$ of the body size of the 3D model:
\begin{eqnarray}
\bold{A}=\bold{A}_{ini}+\mathcal{N}(0,\sigma_{\alpha}^2E),
\end{eqnarray}
where $E$ is the identity matrix. The standard deviation $\sigma_{\alpha}$ used to generate the Gaussian noise is 5 cm for each size parameter. Each hypothesis $\bold{A}$ represents one set of possible values of the body size parameters (Fig. \ref{figure:size}):
\begin{equation}
\bold{A}=[a_1,a_2,\cdots,a_{18}]^T.
\end{equation}
Because of the large number of parameters, the parameters are estimated in a three-step hierarchical way. Size of torso, waist and head in front view $\bold{O}^1$, size of torso, waist and head in side view $\bold{O}^2$, and size of arms in front view $\bold{O}^{3}$ are estimated separately:
\begin{eqnarray}
&\bold{A}=[\bold{O}^1,\bold{O}^2,\bold{O}^{3}]^T,&\bold{O}^1=[a_1,\cdots,a_8]^T,\nonumber\\
&\bold{O}^2=[a_9,\cdots,a_{12}]^T,&\bold{O}^3=[a_{13},\cdots,a_{18}]^T.
\end{eqnarray}
According to the subset $\bold{O}^j,j\in {1,2,3}$, that is being estimated, reference images $I_\textrm{ref}^j$ observing the test person doing corresponding reference postures are captured and used to estimate the size parameters. Reference postures $\bold{d}_\textrm{ref}^j$ of the test person are assumed known. The entire process of offline model initialization is described in Algorithm \ref{alg0}.
\begin{algorithm}
\begin{small}
\caption{Offline Model Initialization}
\label{alg0}
\begin{algorithmic}
\REQUIRE{$\bold{A}_{ini}$.}
\STATE{Generate $M$ hypotheses for estimating size parameters:}
\FOR {$i=1:1:M$}
    \STATE{$\bold{A}_i=\bold{A}_{ini}+\mathcal{N}(0,\sigma_{\alpha}^2E)$;}
\ENDFOR
\STATE{3-step hierarchical estimation of subsets:}
\FOR {$j=1:1:3$}
    \WHILE{Initialization accuracy not reached}
        \STATE{Find the best subset $\tilde{\bold{O}}^{j}$ using corresponding reference posture $\bold{d}_\textrm{ref}^j$ and reference image $I_\textrm{ref}^j$:}
        \STATE{$\tilde{\bold{O}}^{j}=\displaystyle\mathop{\mathrm{argmin}}\limits_{\bold{O}^{j}_i,i=1,\cdots,M}W(\bold{d}_\textrm{ref}^j,I_\textrm{ref}^j)\label{equ101}$;}
        \STATE{Generate $M$ new hypotheses of corresponding subset:}
        \FOR{$i=1:1:M$}
        \STATE{$\bold{O}^{j}_i=\tilde{\bold{O}}^{j}+\mathcal{N}(0,\sigma_{\alpha}^2E)$;}
        \ENDFOR
    \ENDWHILE
\ENDFOR
\end{algorithmic}
\end{small}
\end{algorithm}
\subsection{Online Motion Tracking}
\indent Once the human model is initialized, the body posture parameters of the observed person are continuously estimated in the online motion tracking phase. We apply a hierarchical particle filter to perform online motion tracking as described in Algorithm \ref{alg1}: first, initialize particles for tracking torso, left arm and right arm separately; then resample these particles with DRS, using Algorithm \ref{alg3}. Here, each particle of the particle filter represents a certain human posture.
\begin{algorithm}
\begin{small}
\caption{Online Motion Tacking}
\label{alg1}
\begin{algorithmic}
\REQUIRE{Body size parameters are initialized.}
\STATE{Initialize particles:}
\STATE{Set the index of estimation cycle $k$ to 0, $k=0;$}
\FOR {$i=1:1:N$}
    \STATE{$\bold{d}_i^0=\bold{d}_\textrm{ref}+\mathcal{N}(0,\sigma_{\beta}^2E)$;}
\ENDFOR
\STATE{Assign each particle $\bold{d}_{{i}}^0,i=1,\cdots, N$ a cost using (\ref{equ100});}
\WHILE{New observation image $I_t$}
\STATE{Sample and resample with DRS, as described in Algorithm \ref{alg3};}
\ENDWHILE
\end{algorithmic}
\end{small}
\end{algorithm}
The general idea of our hierarchical particle filter is to divide the whole estimation procedure into three steps that are listed below:
\begin{itemize}
	\item First, estimate the orientation of the torso (3 DOF) and the position of the torso (1 DOF).
    \item Second, estimate the orientation of the left arm (2 DOF for left shoulder, 2 DOF for left elbow).
    \item Third, estimate the orientation of the right arm (2 DOF for right shoulder, 2 DOF for right elbow).
\end{itemize}
Note that the second and the third step can be processed parallel.\\
\indent Particles $\bold{d}_i^0$ of the initialization step are generated by adding Gaussian noise $\mathcal{N}(0,\sigma_{\beta}^2E)$ to the reference posture $\bold{d}_\textrm{ref}$ from which online motion tracking starts:
\begin{equation}
\bold{d}_i^0=\bold{d}_\textrm{ref}+\mathcal{N}(0,\sigma_{\beta}^2E),~i=1,~\cdots,~N,\label{equ0006}
\end{equation}
where $N$ denotes the total number of the used particles. Here, the superscript is the index of estimation cycle which starts with ``0" indicating the initialization. The subscript is the particle index.
\subsection{Sampling and Resampling}
\indent In the following, we introduce first a standard particle filter that uses SRS (stratified resampling strategy). After that, we discuss our DRS.
\subsubsection{SRS}
\indent For tracking with SRS, the sampling and resampling part of online motion tracking is executed using SRS, as described in Algorithm \ref{alg4}. In the $k$th estimation cycle of SRS, each particle $\bold{d}_i^k,i=1,~\cdots,N,$ is assigned a probability $p_i^k$:
\begin{equation}\label{equ0001}
p_i^k =\exp\left({-\frac{{W_i^k}^2}{2{\sigma_\gamma}^2}}\right),
\end{equation}
where $W_i^k$ denotes the cost of particle $\bold{d}_i^k$. For the tracking of torso, left arm and right arm, $\sigma_\gamma$ has different values, as listed in Table \ref{TAB:Parameters used for all experiments.}.\\
\indent Having calculated the probability, the normalized probability $\omega_i^k$ for each particle can be obtained as follows:
\begin{eqnarray}
\omega_i^k=\frac{1}{\sum\limits_{i=1}^N p_i^k}p_i^k.\label{equ201}
\end{eqnarray}
Using the normalized probability, we calculate for each particle a cumulative probability as follows:
\begin{eqnarray}
L_i^k=\sum\limits_{j=1}^i\omega_j^k, \label{equ202}
\end{eqnarray}
where $L_i^k$ denotes the cumulative probability for particle $\bold{d}_i^k$ in estimation cycle $k$.
\begin{algorithm}
\begin{small}
\caption{SRS}
\label{alg4}
\begin{algorithmic}
\REQUIRE{Particles are initialized.}
\STATE{$\bold{d}_1^k=\displaystyle\mathop{\mathrm{argmin}}\limits_{\bold{d}_{i}^{k-1},i=1\cdots N}W(\bold{d}_{i}^{k-1},I_t)$;}
\FOR {$i=2:1:N$}
    \STATE{Generate a random number $r$ with the uniform distribution $\mathcal{U}[0,1)$ according to the stratified sampling \cite{16}};
    \STATE{$\bold{d}_i^{k}=\bold{d}_\psi^{k-1}+\mathcal{N}(0,\sigma_{\beta}^2E)$, with $\psi=min\{j|r\leq L_{j}^{k-1}\}$;}
    \STATE{Assign each particle $\bold{d}_i^k, i=1\cdots~ N$ a normalized probability using (\ref{equ201}) and calculate the cumulative probability for each particle using (\ref{equ202});}
\ENDFOR
\STATE{$k$++};
\end{algorithmic}
\end{small}
\end{algorithm}
\subsubsection{DRS}
\begin{figure}[bp]
    \centering
    \includegraphics[height=8cm]{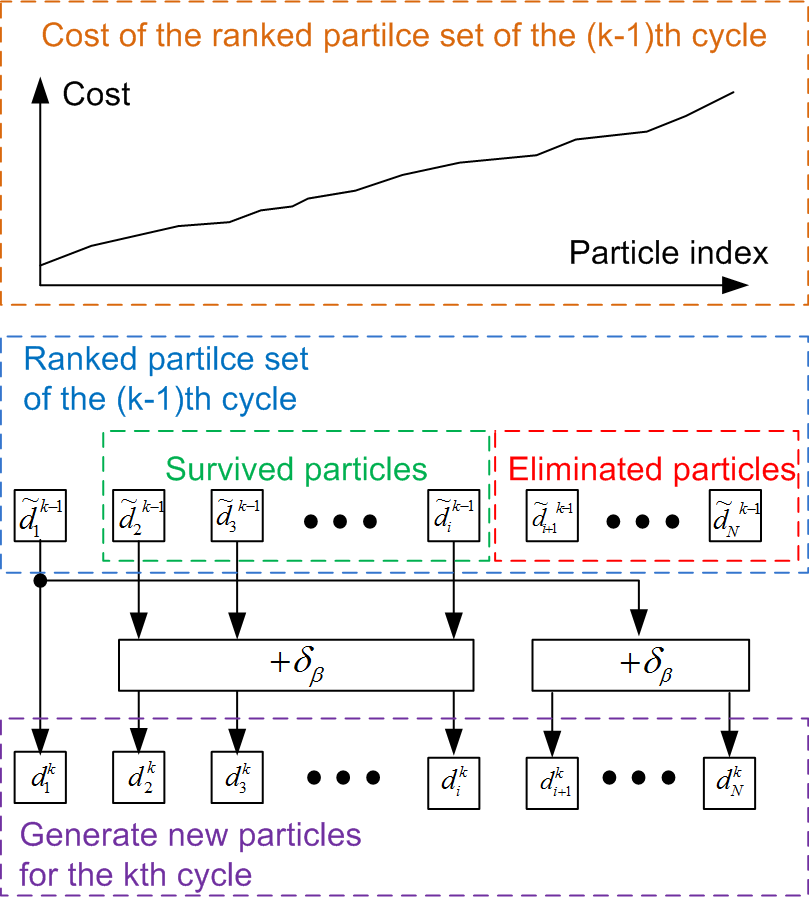}
    \caption{Sample and resample with DRS.}
    \label{FIG:Sample and resample with DRS}
\end{figure}

\indent The standard particle filter with SRS uses a Gaussian distribution for the observation probability density, and it samples in a total stochastic way, thus, the selectivity of particles degrades as the modeling error of the observation model increases. This situation occurs especially when the physical model cannot perfectly match the appearance of the human body. Our DRS uses a survival of the fittest paradigm which ensures the selection of the best hypothesis. Therefore, it is especially suitable for filters with a low number of hypotheses. Moreover, it allows to directly process error values without a non-linear transformation by a Gaussian pdf.\\
\indent Our DRS generates new particles according to a survival-rate $e$, as described in Algorithm \ref{alg3}. Let $\tilde{\bold{d}}^{k-1}_i$ denote the ranked list of posture parameter with $W(\tilde{\bold{d}}_{i}^{k-1},I_t)<W(\tilde{\bold{d}}_{i+1}^{k-1},I_t)$. The first particle $\bold{d}_1^k$ in the $k$th estimation cycle is the same as the best particle in the $(k-1)$th estimation cycle $\tilde{\bold{d}}^{k-1}_1$:
\begin{equation}
\bold{d}_1^k=\tilde{\bold{d}}^{k-1}_1=\displaystyle\mathop{\mathrm{argmin}}\limits_{\bold{d}_{i}^{k-1},i=1\cdots N}W(\bold{d}_{i}^{k-1},I_t).\label{equ0002}
\end{equation}
Multiplying the survival-rate $e$ with the total number of particles $N$ gives the amount of particles $e\cdot N$ that survive the $(k-1)$th estimation cycle. The survived particles are added with a random Gaussian noise vector $\mathcal{N}(0,\sigma_{\beta}^2E)$ so as to produce new particles:
\begin{equation}
\bold{d}_i^k=\tilde{\bold{d}}^{k-1}_i+\mathcal{N}(0,\sigma_{\beta}^2E),~1<i\leq e\cdot N.\label{equ0003}
\end{equation}
In the $k$th estimation cycle, the particles that did not survive are replaced by particles drawn from the neighbourhood of the best particle:
\begin{equation}
\bold{d}_i^k=\tilde{\bold{d}}^{k-1}_1+\mathcal{N}(0,\sigma_{\beta}^2E),~e\cdot N<i\leq N.\label{equ0004}
\end{equation}
The particle with the lowest cost is considered to be the estimation result for each estimation cycle. The process of DRS is illustrated in Fig. \ref{FIG:Sample and resample with DRS}.
\begin{algorithm}
\begin{small}
\caption{DRS}
\label{alg3}
\begin{algorithmic}
\REQUIRE{Particles are initialized.}
\STATE{Rank particles $\bold{d}_{{i}}^{k-1},i=1,\cdots, N$, according to their costs, then store the ranked particles into the particle list $\tilde{\bold{d}}^{k-1}_i$};
\FOR {$i=1:1:N$}
    \IF{$i=1$}
        \STATE{$\bold{d}_1^k=\tilde{\bold{d}}^{k-1}_1$;}
    \ELSIF{$1<i\leq e\cdot N$}
        \STATE{$\bold{d}_i^k=\tilde{\bold{d}}^{k-1}_i+\mathcal{N}(0,\sigma_{\beta}^2E)$; }
    \ELSE
        \STATE{$\bold{d}_i^k=\tilde{\bold{d}}^{k-1}_1+\mathcal{N}(0,\sigma_{\beta}^2E)$;}
    \ENDIF
    \STATE{Assign each particle a cost using (\ref{equ100});}
\ENDFOR
\STATE{$k=k+1$};
\end{algorithmic}
\end{small}
\end{algorithm}
\section{EXPERIMENTS AND DISCUSSIONS}\label{4}
\begin{figure}[tp]
    \centering
    \includegraphics[height=4.8cm,width=8cm]{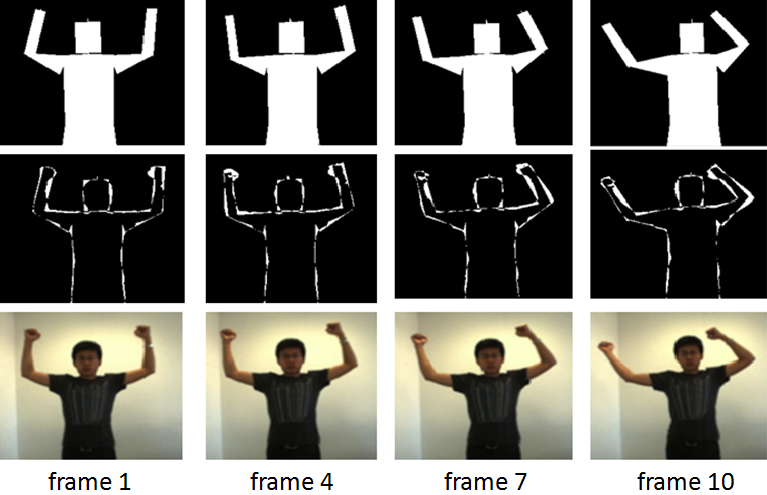}
    \caption{Online motion tracking with DRS: wave hands. First row: the best estimate; second row: matching result; third row: raw image.}
    \label{FIG:Online motion tracking: waving hands.}
\end{figure}

\begin{figure}[tp]
    \centering
    \includegraphics[height=4.8cm,width=8cm]{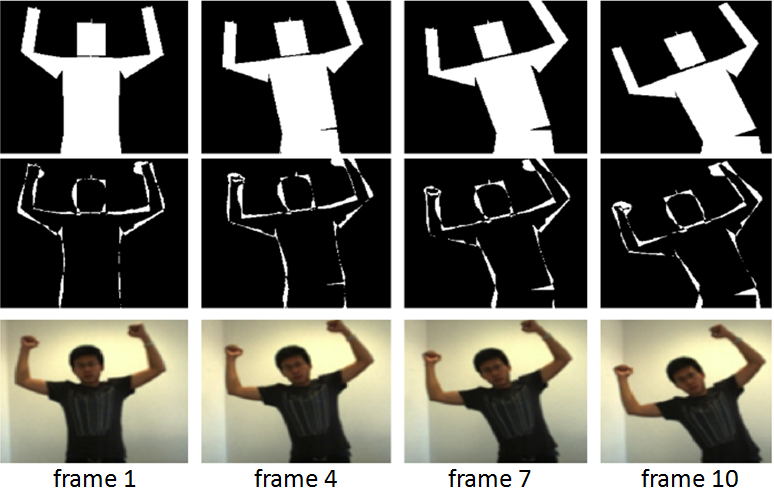}
    \caption{Online motion tracking with DRS: bend aside. First row: the best estimate; second row: matching result; third row: raw image.}
    \label{FIG:Online motion tracking: bend aside.}
\end{figure}

\begin{figure}[tp]
    \centering
    \includegraphics[height=4.8cm,width=8cm]{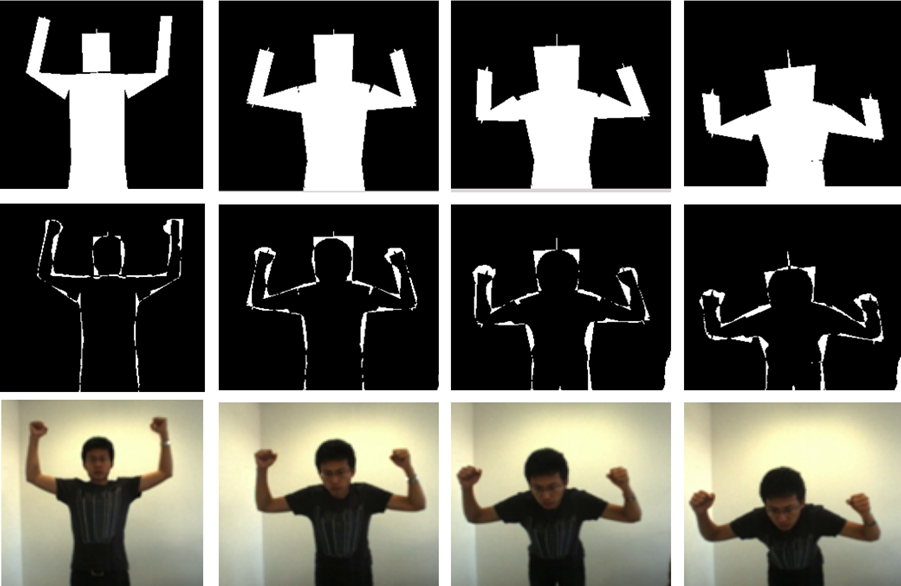}
    \caption{Online motion tracking with DRS: bow forward. First row: the best estimate; second row: matching result; third row: raw image.}
    \label{FIG:Online motion tracking: Bowing forward.}
\end{figure}

\indent The experiments are performed with a BumbleBee Stereo Camera BB-COL-20, whereas only one camera of the two is used. A quad-core AMD Phenom$^{\textrm{TM}}$ II X4 945 (3.0 GHz) processor is used as CPU. However, the current implementation is single-threaded. The ``Sequential Monte Carlo Template Class" (SMCTC) \cite{60} library in C++ is used as the framework for implementing particle filter. The OpenCV library \cite{03} is used for image processing and camera calibration. The resolution of input images is 320$\times$240. \\
\indent For the choice of the survival-rate $e$, imagine two extreme situations: just the best particle survives ($e\approx0$) and all the particles survive ($e=1$). In the case of $e\approx0$, the procedure becomes a greedy local optimization, which redistributes the particles just around the best one. If the observed human motion is fast, the estimation tends to get lost due to the limited search range, which is decided by the noise used to generate new particles. In the case of $e=1$, the search range of the particle set is much bigger than that in local optimization, but the estimation performance is rather rough, because in this case the resampling is blocked, so that all the particles are just distributed randomly in the whole sample space. Thus, the reasonable choice of $e$ should be a compromise of these two cases, which ensures a big search range and a satisfactory accuracy. The values of used parameters are summarized in Table \ref{TAB:Parameters used for all experiments.}.\\
\begin{table}[bp]
\caption{Parameters used for all experiments.}
\label{TAB:Parameters used for all experiments.}
\centering
    \begin{tabular}{|c||c|c|c|}  \hline
        &Torso&Left arm& Right arm\\ \hline
    $N$&200&200&200\\ \hline
    $e$&0.2&0.4&0.4\\ \hline
    $\sigma_{\beta}$ for rotation (radian)&0.1&0.25&0.25\\ \hline
    $\sigma_{\beta}$ for translation (cm)&2&-&-\\ \hline
    $\sigma_{\gamma}$ (pixel)&8000&4000&4000 \\ \hline
    \end{tabular}
\end{table}
\indent Motion used in the following experiments consists of a sequence of three fundamental motions: ``wave hands" (Fig. \ref{FIG:Online motion tracking: waving hands.}), ``bend aside" (Fig. \ref{FIG:Online motion tracking: bend aside.}) and ``bow forward" (Fig. \ref{FIG:Online motion tracking: Bowing forward.}). These figures show examples of online motion tracking using 600 particles and DRS. As can be seen, the proposed system can stably track human motions without self-occlusion, using 600 particles ($N=600$). With 600 particles, we achieve real-time tracking in 15 fps.
\subsection{Ground Truth Evaluation}
\begin{figure}[tp]
   \centering
      \includegraphics[height=4cm]{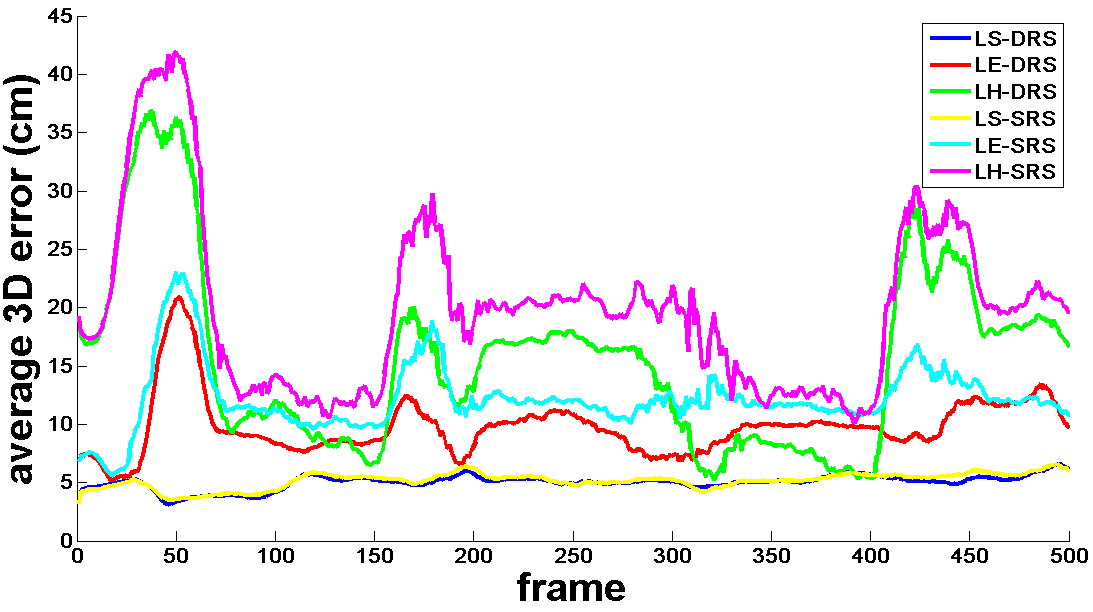}
    \caption{Average 3D error over 100 times: left shoulder (LS), left elbow (LE) and left hand (LH).}
   \label{figure:GT1}
   \end{figure}

\begin{figure}[tp]
   \centering
      \includegraphics[height=4cm]{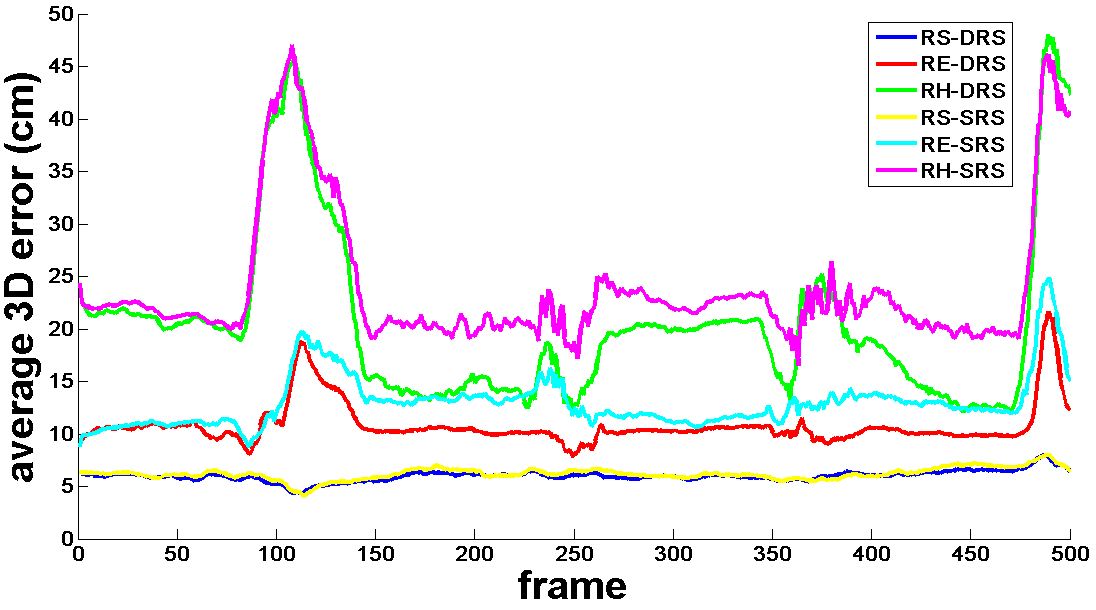}
    \caption{Average 3D error over 100 times: right shoulder (RS), right elbow (RE) and right hand (RH).}
   \label{figure:GT2}
   \end{figure}

\indent The Moven inertial motion capture suit uses 17 inertial motion trackers and an advanced articulated human model to track human motion. It is utilized to measure the underlying body posture parameters that is considered to be the ground truth of experiments. The Moven system provides Cartesian positions of the body segments. For the comparison, the tracking result in the joint space of the proposed method is converted to Cartesian position of corresponding body parts by forward kinematics using the estimated body size parameters. Here, 3D positions of shoulder, elbow and hand are compared. It should be mentioned that we could not achieve accurate comparison of both systems but rough comparison only. The reason is that different human kinematic models are used in both systems and the real positions of corresponding body parts are not defined in the same way. In the future, an accurate measure for quantitative comparison will be investigated.\\
\indent The systems are synchronized with the image captured by the camera. Human motion tracking is repeated 100 times on a video of example video, using the proposed system. As tracking error, we measure the average 3D distance between the corresponding body parts of the Moven and the proposed system in these 100 times. The results are shown in Fig. \ref{figure:GT1} and Fig. \ref{figure:GT2}. DRS achieves better 3D tracking accuracy than SRS in hand and elbow tracking. The accuracy for tracking hand is worse than the accuracy for tracking elbows and shoulders, because the hand of the Moven human model and that of the proposed human model are differently modeled. The hand in Moven system are located at wrists, whereas the hand in the proposed system are at the end of fists. This difference results in an extra error of 10 cm approximately for hand tracking. Taking into account that the tracking is only based on monocular silhouette-matching, and considering the difference of both systems, the proposed system shows good 3D tracking performance.
\subsection{2D Error and Reproducibility of the System Output}
\indent The system output of each estimation cycle refers to the best estimate (particle) that has the lowest cost. In the following, we compare DRS with SRS regarding two aspects of the system output: 2D error and reproducibility.
\subsubsection{2D error}
\begin{figure}[tp]
    \centering
    \includegraphics[height=4.3cm]{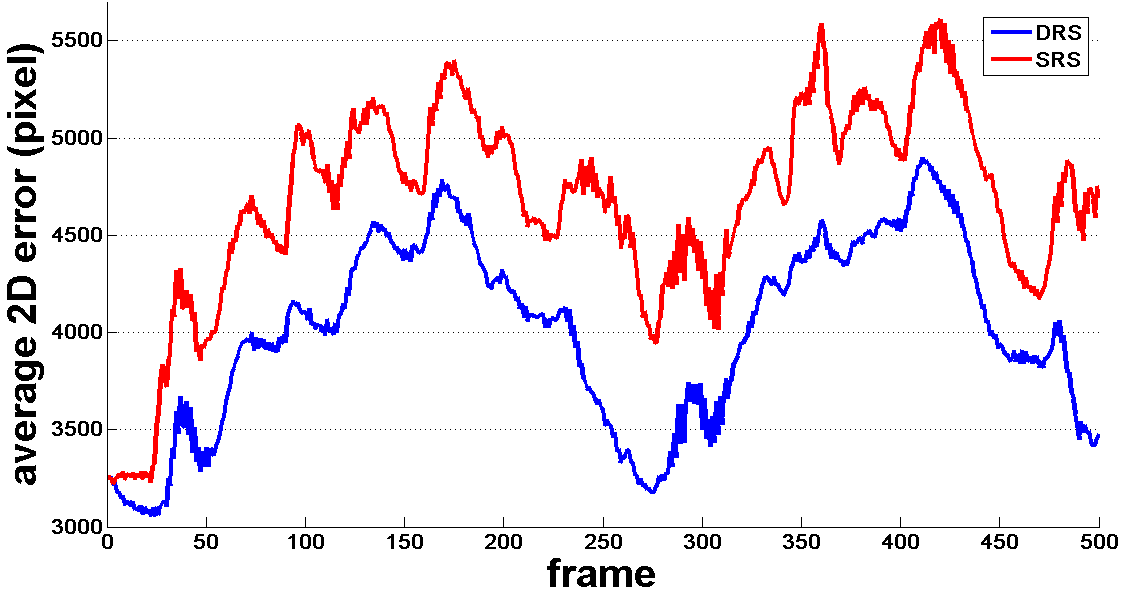}
    \caption{Average 2D error over 100 runs: DRS outperforms SRS, because system output obtained by DRS has less cost for each input image.}
    \label{FIG:Distinction of the best estimate: DRS outperforms SRS.}
\end{figure}

\indent 2D error of the system output is represented by its cost. In this experiment, we measure the mean of the cost of the system output over 100 times and we show the result in Fig. \ref{FIG:Distinction of the best estimate: DRS outperforms SRS.}. Here, we conclude that DRS outperforms SRS with respect to 2D error, because the system output obtained by DRS has less cost for each input image, although the same number of particles are used in both cases. Since we track human motion just based on silhouette-matching, the DRS is a better choice compared with SRS.
\subsubsection{Reproducibility}
\indent Ideally, we expect that the system output should show the same performance on the same input. This feature is important for some applications, such as imitation learning of a humanoid robot. Here, we test the reproducibility of the system output by measuring the mean and standard deviation of each DOF ($d_i,i=1,\cdots,12$) of the system output over 100 times. Experiment shows that tracking with DRS provides better reproducibility of the system output, because the standard deviation of the system output obtained by DRS is smaller than that obtained by SRS. One example of the results is shown in Fig. \ref{FIG:Reproducibility test: one rotational parameter of left elbow, dashed lines are standard deviation lines.}, in which one rotational DOF of left elbow is demonstrated. Here, we can see that the mean value (solid lines) is similar in both cases, but the standard deviation (dashed lines) of the system output obtained by DRS is smaller throughout the entire tracking. This result indicates that the system output estimated by DRS is better reproducible than that by SRS.\\
\indent The DRS works in a survival of the fittest paradigm, and it contains no non-linear processing of the error value, thus, it is more reliable than the SRS for tracking with a small number of particles (600 particles for 12 dimensions).
\begin{figure}[tp]
    \centering
    \includegraphics[height=4.3cm]{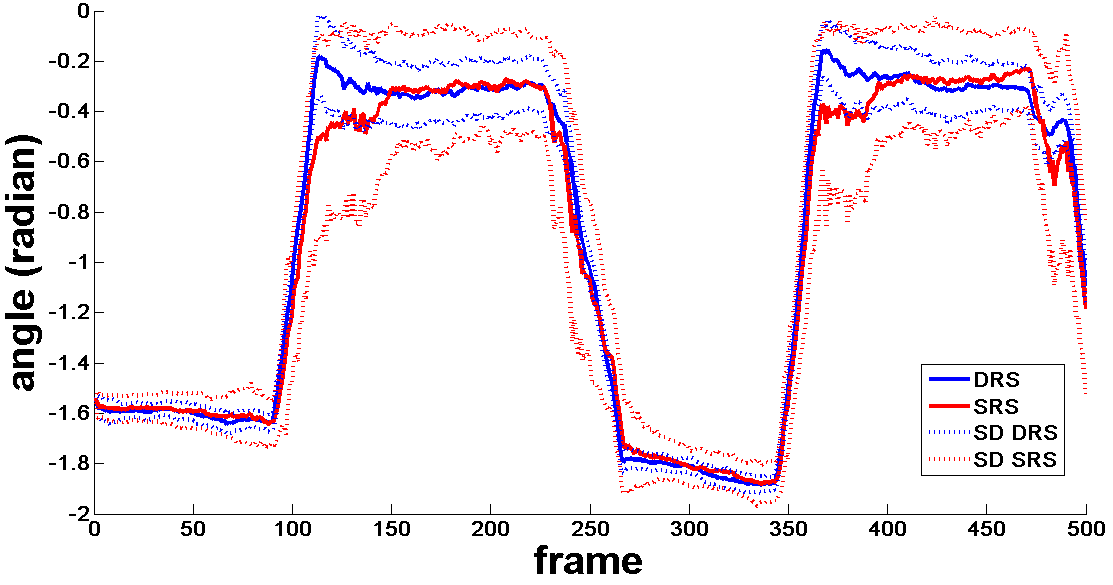}
    \caption{Reproducibility test: one rotational parameter of left elbow, dashed lines are corresponding standard deviation (SD) lines.}
    \label{FIG:Reproducibility test: one rotational parameter of left elbow, dashed lines are standard deviation lines.}
\end{figure}


\section{CONCLUSIONS}
\indent This paper presents an upper body motion capture system based on monocular silhouette-matching, which is built on the basis of a hierarchical particle filter. A novel deterministic resampling strategy (DRS) is applied so as to explore the search space efficiently. Using the ground truth data collected by Moven system, the tracking performance is quantitatively evaluated. As experiments show, the DRS outperforms SRS regarding the 3D and 2D tracking error with the same number of particles. Moreover, reproducibility test show that tracking results of DRS are more stable compared to SRS. \\
\indent Additionally, a new 3D articulated human upper body model with the name 3D cardbox model is created and is proven to work successfully for motion tracking. Compared with the 3D volumetric model, our model has less planes to project while rendering the human model image, which leads to less computational cost. The proposed system achieves a stable real-time tracking of human motion in 15 fps, using 600 particles. It is worth mentioning that motions towards the camera, such as ``bow forward", is also well tracked by the proposed system.\\
\indent Currently, the proposed system tracks human motion just through silhouette-matching. In the future work, we will incorporate more sophisticated cues so as to handle motions with self-occlusion.




\begin{thebibliography}{99}
\bibitem{00} R. Poppe, ``Vision-based human motion analysis: An overview", \emph{Computer Vision and Image Understanding}, 2007, pp. 4-18.

\bibitem{1} P. Azad et.al., ``Stereo-based Markerless Human Motion Capture for Humanoid Robot Systems", \emph{Proc. of IEEE International Conference on Robotics and Automation}, 2007, pp. 3951-3956.
\bibitem{1-1} P. Azad, T. Asfour and R. Dillmann, ``Robust Real-time Stereo-based Markerless Human Motion Capture", \emph{IEEE-RAS International Conference on Humanoid Robots}, 2008, pp. 700-707.
\bibitem{4} J. Deutscher, A. Blake, and I. Reid, ``Articulated Body Motion Capture by Annealed Particle Filtering", \emph{International Conference on Computer Vision and Pattern Recognition (CVPR)}, 2000, pp. 2126-2133.
\bibitem{4-1} J. Deutscher, A. Davison, and I. Reid, ``Automatic partitioning of high dimensional search spaces associated with articulated body motion capture", \emph{International Conference on Computer Vision and Pattern Recognition}, 2001, pp. 669-676.
\bibitem{4-2} M. A. Brubaker and D. Fleet, ``The kneed walker for human pose tracking", \emph{International Conference on Computer Vision and Pattern Recognition (CVPR)}, 2008, pp. 1-8.


\bibitem{5} L. Sigal, S. Bhatia, S. Roth, M. Black, and M. Isard, ``Tracking loose-limbed people", \emph{Proc. of IEEE International Conference on Computer Vision and Pattern Recognition}, 2004, pp. 421-428.
\bibitem{7-1} J. Gall, B. Rosenhahn, T. Brox, and H. P. Seidel, ``Optimization and filtering for human motion capture", \emph{International Journal of Computer Vision}, 2008, pp. 75-92.
\bibitem{14} S. Knoop, S. Vacek, and R. Dillmann, ``Modeling Joint Constraints for an Articulated 3D Human Body Model with Artificial Correspondences in ICP", \emph{International Conference on Humanoid Robots (Humanoids)}, 2005, pp. 74-79.

\bibitem{15} \url{http://www.openni.org}.
\bibitem{15-1} Z. Li and D. Kuli$\acute{\text{c}}$, ``Paritcle Filter Based Human Motion Tracking", \emph{International Conference on Control, Automation, Robotics and Vision}, 2010, pp. 555-560.
\bibitem{16} G. Kitagawa, ``Monte Carlo filter and smoother for non-Gaussian nonlinear state space models", \emph{Journal of Computational and Graphical Statistics}, vol. 5, no. 1, pp. 1-25, 1996.


\bibitem{24} Y. Huang, T. S. Huang, ``Model-based human body tracking", \emph{Proceedings of the International Conference on Pattern Recognition (ICPR��02)}, 2002, pp. 552-555.

\bibitem{33} R. Kehl, L. V. Gool, ``Markerless tracking of complex human motions from multiple views", \emph{Computer Vision and Image Understanding (CVIU)}, vol. 104, no. 2, pp. 190-209, 2006.

\bibitem{34} J. Carranza, C. Theobalt, M. A. Magnor, H. P. Seidel, ``Free-viewpoint video of human actors", \emph{ACM Transactions on Computer Graphics} vol. 22, no. 3, pp. 569-577, 2003.

\bibitem{60} A. M. Johansen, ``SMCTC: Sequential Monte Carlo in C++". \emph{Journal of Statistical Software}, vol. 30, no. 6, pp. 1-41, 2009.
\bibitem{03} G. Bradski, A. Kaehler, {\it Learning OpenCV}. O'Reilly Media, Inc., 2008.

\bibitem{04} Xsens Technologies B.V., ``Moven User Manual". 2008.


\end{thebibliography}
\end{document}